\documentclass[10pt,twocolumn,letterpaper]{article}

\usepackage{cvpr}              

\definecolor{cvprblue}{rgb}{0.21,0.49,0.74}
\usepackage[pagebackref,breaklinks,colorlinks,allcolors=cvprblue]{hyperref}
\usepackage{multirow}

\usepackage{amsmath}
\usepackage{multirow}
\usepackage{bbm}
\usepackage{graphicx}
\usepackage{amssymb}
\usepackage{booktabs}
\usepackage{lscape}
\usepackage{pifont}
\usepackage{algorithm}
\usepackage{caption}
\usepackage{subcaption}
\usepackage{CJKutf8}
\usepackage{makecell}
\usepackage{xcolor}
\usepackage{svg}
\usepackage{listings}
\usepackage{xcolor}

\def\paperID{*****-} 
\def\confName{CVPR}
\def\confYear{2026}

\newcommand\methodname{\textcolor{black}{\textsc{M4-RAG}}}

\title{$\methodname$: A Massive-Scale Multilingual Multi-Cultural Multimodal RAG}

\author{David Anugraha$^1$\thanks{Corresponding author: \url{david.anugraha@stanford.edu}} \text{}, Patrick Amadeus Irawan$^2$, Anshul Singh$^3$,\\
En-Shiun Annie Lee$^{4,5}$, Genta Indra Winata$^6$ \\
$^1$Stanford University$\quad$$^2$MBZUAI\quad$^3$Indian Institute of Science$\quad$$^4$Ontario Tech University\\
$^5$University of Toronto$\quad^6$Capital One\\
}

\makeatletter
\newcommand\mytopfigure{%
  \begin{center}%
    \includegraphics[width=0.95\linewidth]{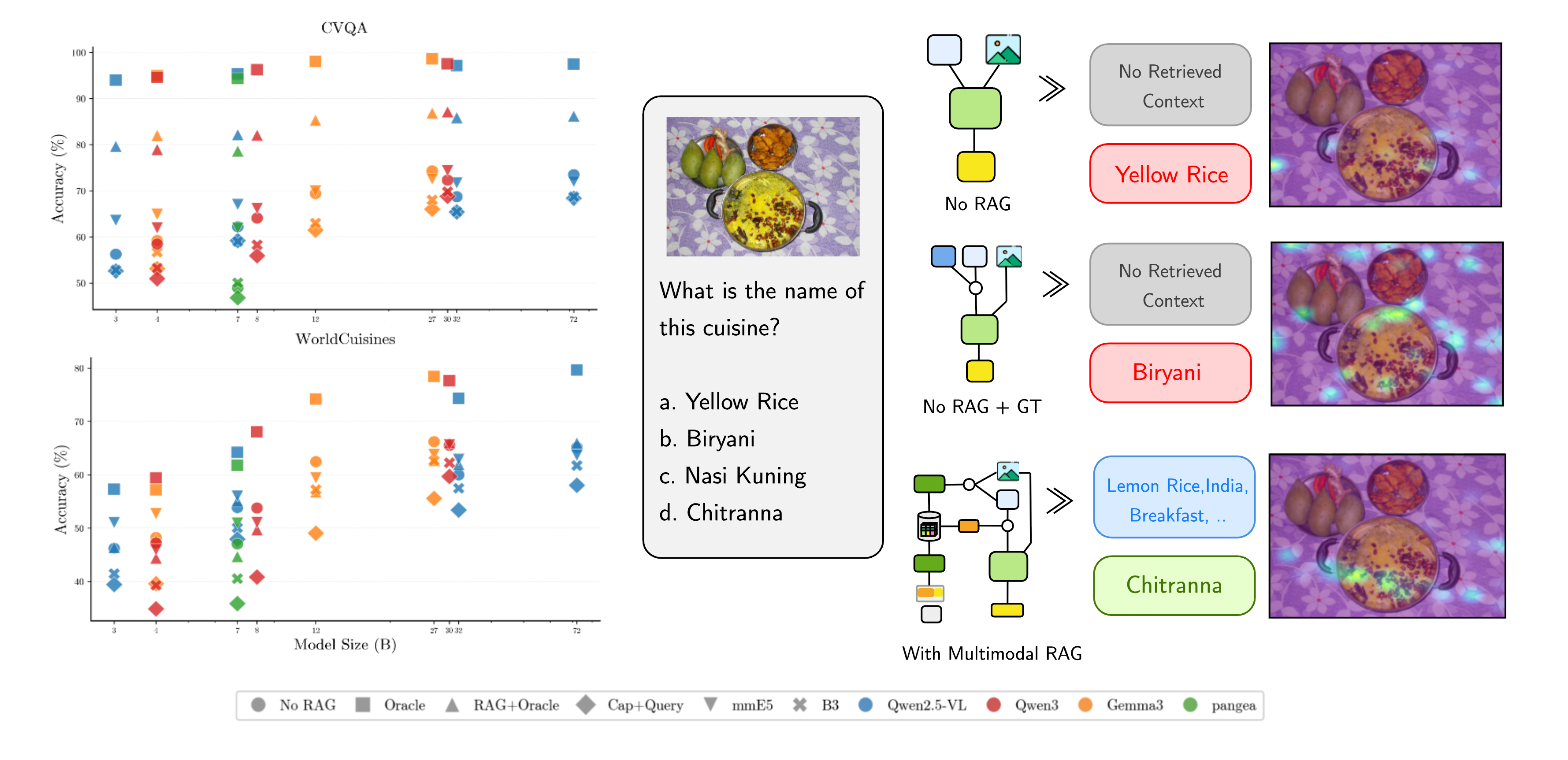}%
    \captionof{figure}{$\methodname$ evaluates state-of-the-art Vision Language Models (VLMs) under culturally grounded retrieval settings. \textbf{Left:} Summarized results on \textsc{CVQA} (top) and \textsc{WorldCuisines} (bottom) across model scales with and without RAG, showing that the retrieved context can either improve or degrade performance depending on the scale. \textbf{Right:} Examples showing how the quality of retrieved context (number of $\bigstar$) affects model behavior and directs attention toward semantically relevant cues. When asked to identify the cuisine in the image, the non-RAG and text-only RAG systems produce incorrect guesses such as ``Yellow Rice'' or ``Biryani''. With multimodal RAG, the system retrieves culturally specific evidence (e.g., Lemon Rice, India, Breakfast), guiding the VLM to the correct answer ``Chitranna.''}
    \label{fig:hookimage}%
  \end{center}%
}
\apptocmd\@maketitle{{\mytopfigure\par}}{}{}
\makeatother

\begin{document}
\maketitle

\begin{abstract}
Vision-language models (VLMs) have achieved strong performance in visual question answering (VQA), yet they remain constrained by static training data. Retrieval-Augmented Generation (RAG) mitigates this limitation by enabling access to up-to-date, culturally grounded, and multilingual information; however, multilingual multimodal RAG remains largely underexplored. We introduce $\methodname$, a massive-scale benchmark spanning 42 languages, 56 regional dialects and registers, and 189 countries, comprising over 80,000 culturally diverse image-question pairs for evaluating retrieval-augmented VQA across languages and modalities. To balance realism with reproducibility, we build a controlled retrieval environment containing millions of carefully curated multilingual documents relevant to the query domains, approximating real-world retrieval conditions while ensuring consistent experimentation. Our systematic evaluation reveals that although RAG consistently benefits smaller VLMs, it fails to scale to larger models and often even degrades their performance, exposing a critical mismatch between model size and current retrieval effectiveness. Our cross-lingual evaluations also reveal significant performance degradation when prompts or retrieved context are provided in non-English languages. The code, datasets, and evaluation protocols
for $\methodname$ are available as open-source at \url{https://github.com/davidanugraha/M4-RAG}.
\end{abstract}

\section{Introduction}
\label{sec:intro}

Recent advances in large language models (LLMs) and vision-language models (VLMs) have demonstrated strong capabilities in reasoning, summarization, and question answering~\cite{yang2024qwen2,dash2025aya,deitke2025molmo,bai2025qwen3,team2025gemma}. However, their reliance on static training corpora often leads to outdated or incomplete knowledge, limiting factual accuracy and cross-domain coverage. Retrieval-Augmented Generation (RAG) addresses this limitation by augmenting model outputs with information retrieved from external knowledge sources~\cite{lewis2020retrieval,winata2024miners}.

RAG has evolved along two complementary directions: multilingual and multimodal RAG. Multilingual RAG~\cite{zhang2023miracl,chirkova2024retrieval, park2025investigating} enables cross-lingual information access, allowing queries and retrieved documents to appear in different languages, while multimodal RAG~\cite{lin2024mm, abootorabi2025ask,faysse2025colpali} incorporates visual or structured inputs such as images, tables, or videos into retrieval and generation pipelines. Despite rapid progress in both areas, their intersection—multilingual multimodal RAG—remains largely unexplored, as shown in Table~\ref{tab:comparison}. This gap is particularly important because real-world knowledge access is inherently both multilingual and multimodal~\cite{thapliyal2022crossmodal}. Cultural knowledge exemplifies this challenge since it is inherently long-tail, region-specific, and not reliably encoded in model parameters even for large models~\cite{myung2024blend,romero2024cvqa,winata2025worldcuisines}, making it an ideal and well-motivated testbed for multilingual multimodal RAG. Consequently, the alignment between cross-lingual retrieval and multimodal representations, the ability of multilingual models to ground information across modalities, and the adequacy of evaluation metrics in capturing these complex dependencies are important, despite remaining underexplored.

To address these challenges, we present $\methodname$, a comprehensive evaluation framework for multilingual, multicultural, and multimodal RAG. Our evaluation spans multiple languages and modalities, covering both text-text and text-image retrieval scenarios. Figure~\ref{fig:hookimage} illustrates how multimodal retrieval can guide model attention toward relevant visual regions to correctly identify the dish as ``Chitranna,'' whereas in the first two settings, it fails to help the VLMs to focus on the correct objects. Through systematic experiments across diverse model families and language configurations, we show that current RAG systems are less effective on larger VLMs and degrade substantially when queries and retrieved contexts differ in language or modality, highlighting limitations in cross-lingual alignment, retrieval robustness, and multimodal reasoning. 

\noindent Our key contributions are summarized as follows:
\begin{itemize}
    \item We introduce the first massive-scale evaluation framework for multilingual multimodal RAG, spanning 42 languages with 56 regional dialects and registers across two cultural VQA datasets: \textsc{CVQA}~\cite{romero2024cvqa} and \textsc{WorldCuisines}~\cite{winata2025worldcuisines}. Table~\ref{tab:comparison} compares our framework with existing multilingual and multimodal datasets.\footnote{The dataset is available at~\url{https://huggingface.co/datasets/davidanugraha/M4-RAG}, and the codebase is released at~\url{https://github.com/davidanugraha/M4-RAG}.}
    \item We conduct a systematic study of retrieval strategies for VLM-based RAG. We find that naive text-based retrieval can degrade performance, while multimodal retrieval provides more reliable gains but does not consistently scale with model size. Furthermore, retrieval relevance correlates with performance but does not guarantee successful evidence integration, particularly for larger models that are less likely to incorporate corrective evidence.
    \item We perform cross-lingual evaluation across 42 languages and find that current VLMs exhibit significant performance degradation when prompts or retrieved context are provided in non-English languages.
\end{itemize}

\noindent We hope this work provides a foundation for developing RAG systems capable of reasoning seamlessly across languages, modalities, and cultures.

\begin{table*}[!th]
\centering
\resizebox{\textwidth}{!}{
    \begin{tabular}{lcccccccc}
    \toprule
    \textbf{Dataset} & \textbf{\# Size} & \textbf{\# Languages} & \textbf{\# Dialects} & \textbf{Modality} & \textbf{Domains} & \textbf{Retrieval Source} & \textbf{License} \\
    \midrule
    TyDiQA~\cite{clark2020tydi} & 204k & 11 & N/A & Text & Multiple$^\dagger$ & Wikipedia & N/A \\
    MLQA~\cite{lewis2020mlqa} & 46k & 7 & N/A & Text & Multiple$^\dagger$ & Wikipedia & CC-BY-SA 3.0 \\
    XOR QA~\cite{asai2021xor} & 40k & 7 & N/A & Text & Multiple$^\dagger$ & Wikipedia & CC BY-SA 4.0 \\
    MKQA~\cite{longpre2021mkqa} & 260k & 26 & N/A & Text& Multiple$^\dagger$ & Wikidata & CC-BY-SA 3.0 \\
    Mintaka~\cite{sen2022mintaka} & 20k & 9 & N/A & Text & Movies, Music, Sports & Wikidata & CC-BY-4.0 \\
    & & & & & Books, Geography, Politics \\
    & & & & & Video Games, History \\ 
    MIRACL~\cite{zhang2023miracl} & 79k & 18 & N/A & Text & Multiple$^\dagger$ & Wikipedia & N/A \\
    AfriQA~\cite{ogundepo2023afriqa} & 12k & 10 & N/A & Text & Multiple$^\dagger$ & Wikipedia & CC-BY-SA 4.0 \\ 
    MIRAGE-Bench~\cite{thakur2025mirage} & 50k & 7 & N/A & Text & Multiple$^\dagger$ & Wikipedia & N/A 
    \\ \midrule
    ViQuAE~\cite{lerner2022viquae} & 3.6k & 1 & N/A & Text, Image & Multiple$^\dagger$ & Wikipedia & N/A \\
    Encyclopedic VQA~\cite{mensink2023encyclopedic} & 221k & 1 & N/A & Text, Image & Fine-grained species, Landmarks & Wikipedia & N/A \\
    \citet{xue2024enhanced} & 500 & 1 & N/A & Text, Image & Genome, Urban & N/A & N/A \\
    UniFashion~\cite{zhao2024unifashion} & 260k & 1 & N/A & Text, Image & Fashion & N/A & N/A \\
    MRAG-Bench~\cite{hu2024mragbench} & 1,353 & 1 & N/A & Text, Image & Multiple$^\dagger$ & Wikipedia, ImageNet~\cite{deng2009imagenet}, & N/A \\ 
    & & & & & & Flowers102~\cite{nilsback2008automated}, StanfordCars~\cite{krause20133d} \\
    Chart-MRAG Bench~\cite{yang2025benchmarking} & 4,738 & 1 & N/A & Text, Image & Family, Race, Politics, Religion, & pewresearch & N/A \\
    & & & & & Economy, International Affairs, \\
    & & & & &  Internet, Scientific Research \\ \midrule 
    $\textbf{\methodname}$ & 80k & 42 & 56 & Text, Image & Vehicles, Food, People & Wikidata, Wikipedia & CC-BY-SA 4.0 \\
    & & & & & Sports, Plants \& Animals, Objects\\
    & & & & & Brands,  Geography, Tradition, Pop Culture \\
    \bottomrule
    \end{tabular}
}
\caption{Comparison of multilingual and multimodal RAG datasets. $\methodname$ offers broader linguistic coverage, spanning 42 languages, and explicitly incorporates regional dialects to provide a more fine‑grained view of dialectal representation. This enables more precise analysis of cultural and linguistic variation. In addition, our benchmark is released under a permissive open‑source license to facilitate reuse and further research. $^\dagger$Details for these entries are not specified in the original papers.}
\label{tab:comparison}
\end{table*}

\section{\methodname}
\label{sec:methodology}

\subsection{Tasks and Objectives}
We propose $\methodname$, an evaluation framework for multilingual multimodal RAG that prioritizes end-to-end task performance while enabling systematic investigation of when and why retrieval systems help or hinder generation quality. Unlike prior work that evaluates retrieval and generation in isolation, we assess their interaction in realistic multilingual multimodal settings, where questions, images, and knowledge sources may span diverse languages.

Formally, given a VQA instance consisting of a question $q$ in language $\ell_q$, an associated image $I$, and a ground-truth response $r^*$, along with a multilingual document corpus $\mathcal{C}$ containing relevant factual and cultural knowledge, the system must produce a response $r$. The RAG pipeline operates in two stages. First, a retriever $R_\theta$ selects the top-$k$ most relevant passages from the corpus:
\begin{equation}
    R_\theta(q, I, \mathcal{C}) = D_k = \{d_1, d_2, \ldots, d_k\},
\end{equation}
where each passage $d_i$ may be in any language $\ell_d$, reflecting real-world retrieval scenarios where relevant information is not restricted to the query language. Then, a VLM $\mathcal{M}$ generates a response conditioned on the question, image, and retrieved context:
\begin{equation}
    \hat{a} = \mathcal{M}(q, I, D_k),
\end{equation}
which is evaluated against $a^*$ using task-specific accuracy metrics.

\subsection{Evaluation Benchmark Source}
We source our VQA pairs from two existing large-scale, culturally-rich datasets, \textsc{CVQA}~\citep{romero2024cvqa} and \textsc{WorldCuisines}~\citep{winata2025worldcuisines}. These benchmarks form the foundation of our evaluation by providing high-quality, human-annotated examples that are both multilingual and deeply grounded in cultural context. Cultural knowledge is particularly well-suited for this evaluation since it is inherently long-tail and region-specific, making it unlikely to be reliably encoded in model parameters even for large models, and thus a natural testbed for retrieval augmentation. In total, these datasets cover 42 languages and 56 regional dialects, with further details provided in the Supplementary Materials.

\paragraph{\textsc{CVQA}.} \textsc{CVQA} is a multilingual dataset with more than 10,000 VQA pairs spanning 10 diverse cultural categories across 30 countries and 31 languages. We include \textsc{CVQA} to complement the domain-specific nature of \textsc{WorldCuisines}, thereby broadening the evaluation landscape with a wider variety of cultural domains and knowledge sources. This diversity is essential for assessing whether RAG systems can retrieve and reason over a broad spectrum of cultural information, rather than performing well only within a single domain.

\paragraph{\textsc{WorldCuisines}.} \textsc{WorldCuisines} is a massive-scale benchmark containing 60k VQA pairs that are parallel across 30 languages and dialects, centered on global cuisine. We select \textsc{WorldCuisines} due to its extensive multilingual parallelism that enables controlled analysis of cross-lingual retrieval behavior under consistent semantic content. The dataset also includes intentionally challenging scenarios, such as adversarial prompts where the provided context is misleading, which offers a valuable stress test to examine whether RAG can help generation models to re-anchor their responses in factual evidence instead.

\begin{figure*}[t]
    \centering
    \begin{subfigure}[t]{0.252\linewidth}
        \centering
        \includegraphics[width=\linewidth]{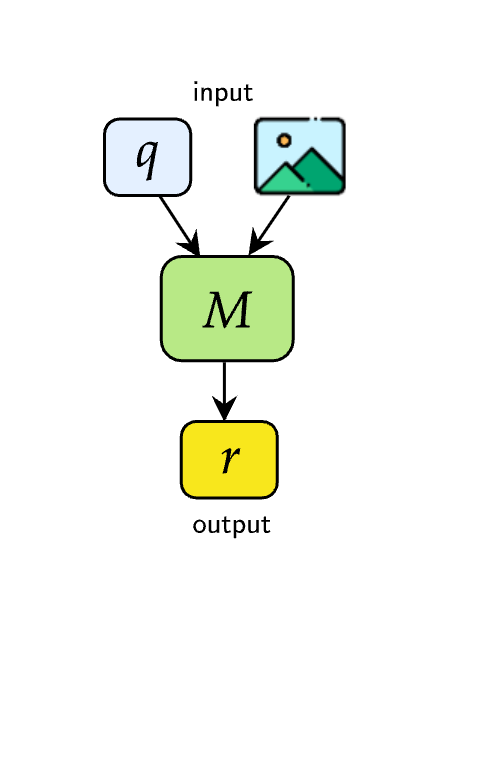}
        \caption{}
        \label{fig:context-by-model}
    \end{subfigure}
    \hfill
    \begin{subfigure}[t]{0.223\linewidth}
        \centering
        \includegraphics[width=\linewidth]{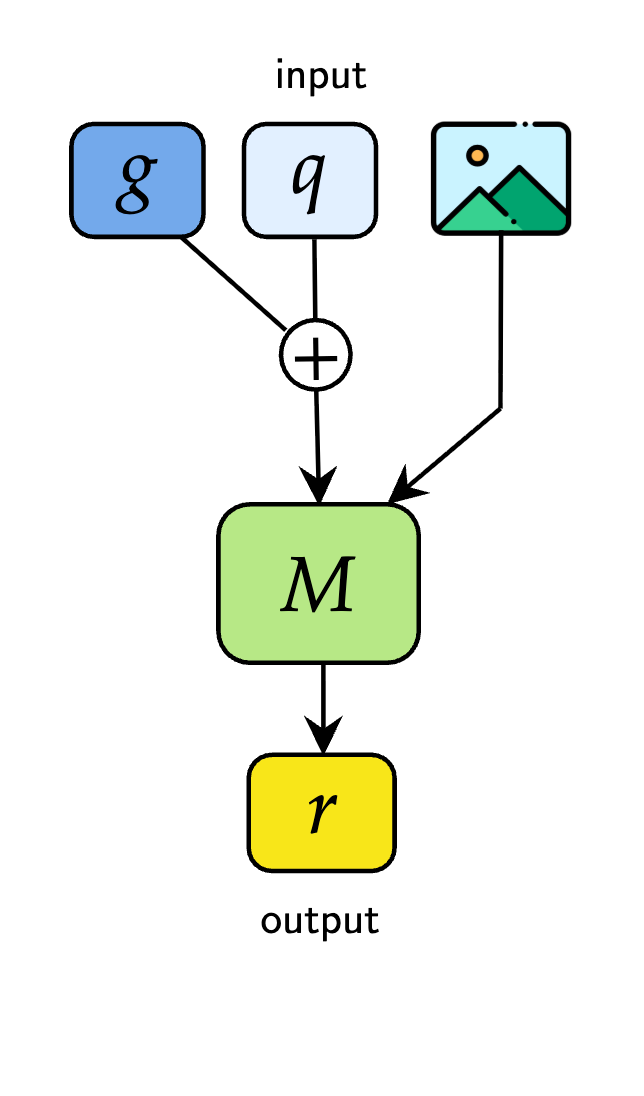}
        \caption{}
        \label{fig:difficulty-country-name}
    \end{subfigure}
    \hfill
    \begin{subfigure}[t]{0.26\linewidth}
        \centering
        \includegraphics[width=\linewidth]{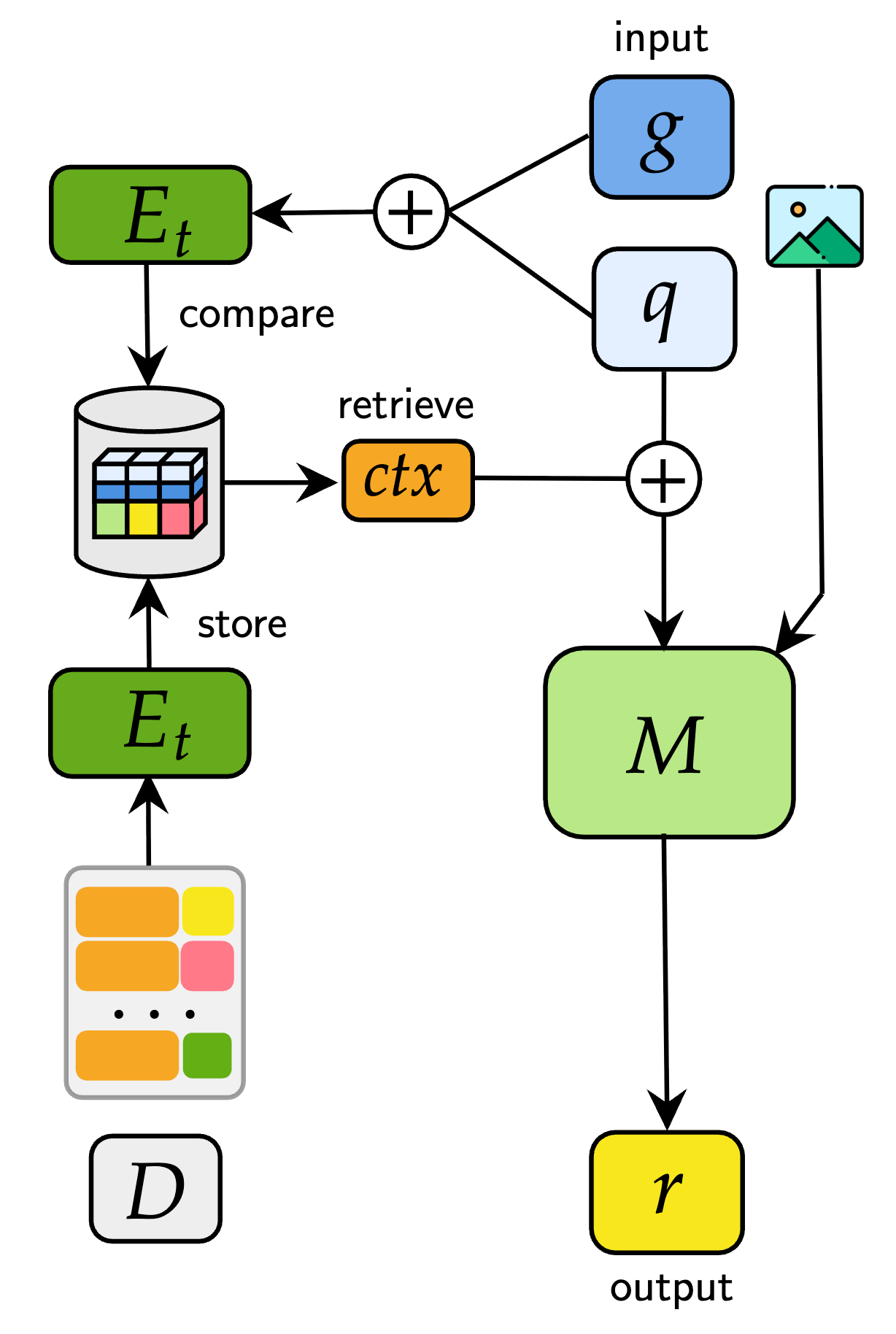}
        \caption{}
        \label{fig:boxplot-geo-desc}
    \end{subfigure}
    \hfill
    \begin{subfigure}[t]{0.245\linewidth}
        \centering
        \includegraphics[width=\linewidth]{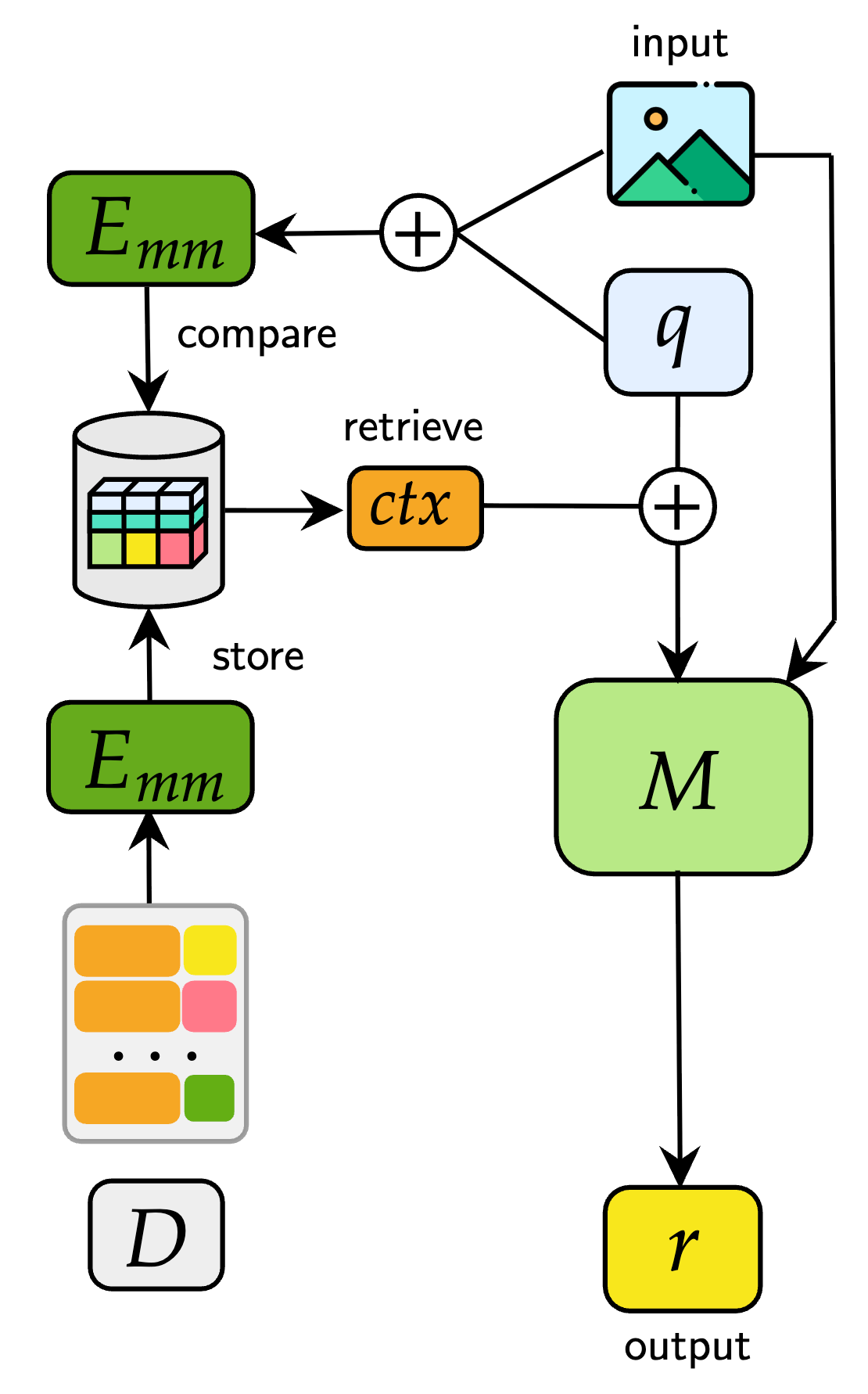}
        \caption{}
        \label{fig:new-fourth-subplot}
    \end{subfigure}
    \caption{The overall evaluation framework of M4‑RAG comprises four configurations: \textbf{(a)} a No‑RAG baseline, where the VLM ($M$) directly takes the question and image as input and predicts a response answer; \textbf{(b)} a No‑RAG setup augmented with oracle context, which is concatenated with the question and image to probe the upper bound of how much perfectly relevant knowledge can help; \textbf{(c)} a text‑based RAG configuration, where a text encoder ($E_{\text{t}}$) encodes the query, compares it against an indexed document collection, retrieves the top textual context, and feeds this retrieved text together with the original inputs; and \textbf{(d)} a multimodal RAG configuration, where documents are stored with embeddings from a text encoder and retrieval can leverage both textual and visual signals with an image encoder ($E_\text{mm}$), yielding richer multimodal context. Across \textbf{(c)} and \textbf{(d)}, the retrieved context is treated as an additional conditioning signal that steers the model toward culturally relevant knowledge while keeping the backbone VLM architecture unchanged.
}
    \label{fig:experiment-figure}
\end{figure*}

\subsection{Knowledge Base Creation}

To systematically evaluate retrieval quality, we construct a new, large-scale multilingual knowledge corpus for each evaluation benchmark. These corpora are built from Wikipedia snapshots dated April 2025 to ensure broad thematic coverage and temporal alignment with the creation timelines of the associated datasets. Another reason to use Wikipedia is for its open licensing and redistribution. This alignment helps preserve contextual fidelity, as cultural information and entity descriptions evolve over time~\citep{creanza2017cultural}.

For each VQA instance, we construct a set of multilingual queries that capture complementary types of evidence. These include a question-only query, an answer-only query, and culturally enriched queries that expand an answer with domain-relevant terms (for example, using “Japanese cuisine” for a sushi-related item). These query types are used in combination to maximize corpus coverage, retrieving the top 25 articles independently in English and in the corresponding target language, to ensure that the non-English passages reflect culturally accurate terminology rather than direct query translations. Articles are parsed into sections using Wikipedia's heading structure, preserving semantic coherence within each retrieval unit, then cleaned to remove non-content elements such as scripts, tables, and navigation elements, and deduplicated across queries and languages, yielding 223,468 articles for \textsc{WorldCuisines} and 306,794 articles for \textsc{CVQA}.

\section{Experimental Setup}
\label{sec:experimental_setup}

\subsection{Retrieval Settings}

To assess the impact of retrieval, we evaluate all VLMs under four main configurations, resulting in six experimental variants per model due to different retrieval strategies. For all RAG-based methods, we retrieve the top-$k$ passages with $k = 5$.

\begin{enumerate}
    \item \textbf{Baseline (No Retrieval):} The VLM receives only the question $q$ and image $I$, without any external context. This serves as a simple baseline to quantify the model's performance without any retrieval assistance.

    \item \textbf{Oracle Context:} The VLM is provided with the oracle context, representing an upper bound on performance. For \textsc{WorldCuisines}, this corresponds to the human-labeled food description from the knowledge base. For \textsc{CVQA}, since no ground-truth evidence passages are provided, we simulate oracle context using a caption generated by \texttt{Qwen2.5-VL-72B-Instruct}, conditioned on the image, question, and human-annotated ground-truth answer. This design ensures the caption is tightly grounded in verified information. To validate caption quality, four annotators evaluated 200 randomly sampled image-caption pairs on a 1-5 Likert scale for relevance to the corresponding question, with all samples receiving a score of 5 with full inter-annotator agreement. Further details are provided in the Supplementary Materials.

    \item \textbf{Text-Based RAG:} Retrieval is performed using textual queries derived from the input, using \texttt{E5}~\citep{wang2024multilingual} as the multilingual dense retrieval. This approach evaluates text-only retrieval by representing visual content through generated captions~\citep{hong2025knowledge, zhao2023retrieving, lin2022retrieval}. Therefore, we consider two settings:
    \begin{itemize}[leftmargin=*]
        \item \textit{Oracle-Query RAG:} The VLM uses the oracle context as the query to retrieve passages. This provides a reliable textual query and serves as a strong reference point for text-based retrieval. 
        \item \textit{Caption-Query RAG:} A caption is first generated from image $I$ and question $q$ using \texttt{Qwen2.5-VL-72B-Instruct}. The VLM then combines the question $q$ and generated caption to retrieve passages, simulating scenario from multiple past works where the image is converted into text for retrieval~\citep{hong2025knowledge, lin2022retrieval}.
    \end{itemize}

    \item \textbf{Multimodal RAG:} The question $q$ and image $I$ are used jointly to retrieve passages, leveraging both textual and visual information holistically. We test two multimodal embedding models: \texttt{mmE5} (11B)~\citep{chen2025mme5} and \texttt{B3} (7B) from VLM2Vec~\citep{jiang2024vlm2vec}.
\end{enumerate}

\begin{figure*}[!th]
  \centering
  \includegraphics[width=\linewidth]{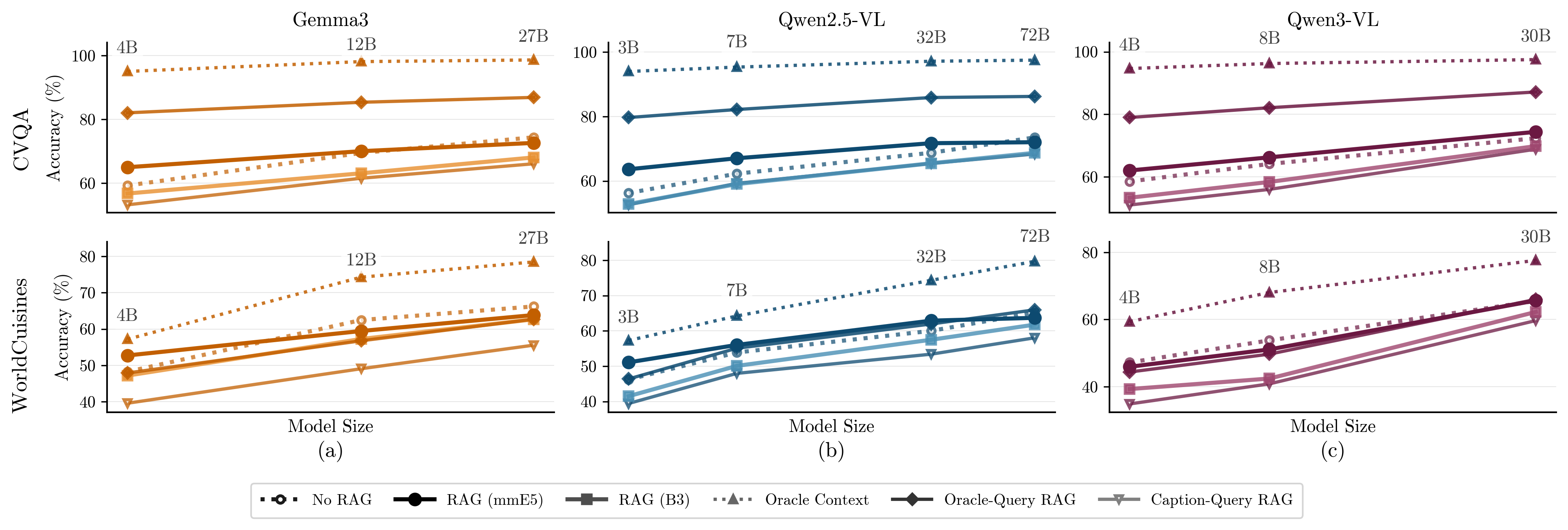}
  \caption{
Overall VQA performance on \textsc{CVQA} and \textsc{WorldCuisines} across different model families and sizes, with different retrieval configurations. Each column corresponds to a VLM family \texttt{(Qwen2.5‑VL, Gemma3, Qwen3-VL)}, and each panel plots accuracy as a function of model size. Across all families and scales, adding retrieval (solid lines) consistently improves over the No‑RAG baseline (dotted black), with the multimodal RAG variants approaching the oracle context upper bound. Gains are especially pronounced on the more culturally nuanced \textsc{WorldCuisines} benchmark, where smaller models with RAG can match or exceed much larger non‑RAG models, illustrating that external knowledge is more beneficial than pure parameter scaling in this setting. Among RAG settings, \texttt{mmE5}‑based retrieval generally outperforms \texttt{B3} and caption‑query retrieval, highlighting the importance of a strong multimodal encoder and joint use of image and query signals to surface culturally relevant evidence.}
  \label{fig:overall-performance}
\end{figure*}

\subsection{Vision Language Models}
We evaluate end-to-end VQA performance across four prominent open-source multilingual VLM families, each available at multiple scales: \texttt{Gemma3}~\citep{team2025gemma} at 4B, 12B, and 27B; \texttt{Qwen2.5-VL}~\citep{bai2025qwen2} at 3B, 7B, 32B, and 72B; \texttt{Qwen3-VL} with reasoning~\citep{yang2025qwen3} at 4B, 8B, and 30B-A3B; and \texttt{Pangea}~\citep{yue2024pangea} at 7B.

\subsection{Multilingual VQA}
We analyze the impact of language on the VQA task in a cross-lingual experimental setting. We measure how the model’s performance changes when the language of its instructional prompts and provided context varies. To do this, we created multilingual versions of our prompts and the oracle contexts for both \textsc{CVQA} and \textsc{WorldCuisines}. We use \texttt{Gemini-2.5-Flash} to produce high-quality translation of two key components. To ensure their fidelity, all translations were subsequently reviewed and validated by annotators.
\begin{itemize}
    \item \textbf{Multilingual Prompts:} The entire instruction template, including system messages and formatting cues, was translated from English into each of the target languages. This creates the \texttt{Multilingual Prompts} setting to see whether models achieve better cultural grounding and task performance when instructions are provided in the native language of the query. We measure this by analyzing the performance change from the English prompt baseline.

    \item \textbf{Multilingual Oracle Context:} Similarly, we created the \texttt{Oracle Multilingual Context} setting to investigate whether models perform better on a cultural VQA task when the evidence is provided in the culture's language. For this oracle setup, we translated the oracle English context into each target language. This allows us to isolate the model's ability to perform cultural reasoning when all information is presented in the target language. By comparing this to the English context baseline, we can directly quantify whether models benefit from language that is aligned with the VQA's cultural context.
\end{itemize}

\begin{figure*}[!th]
  \centering
  \includegraphics[width=\linewidth]{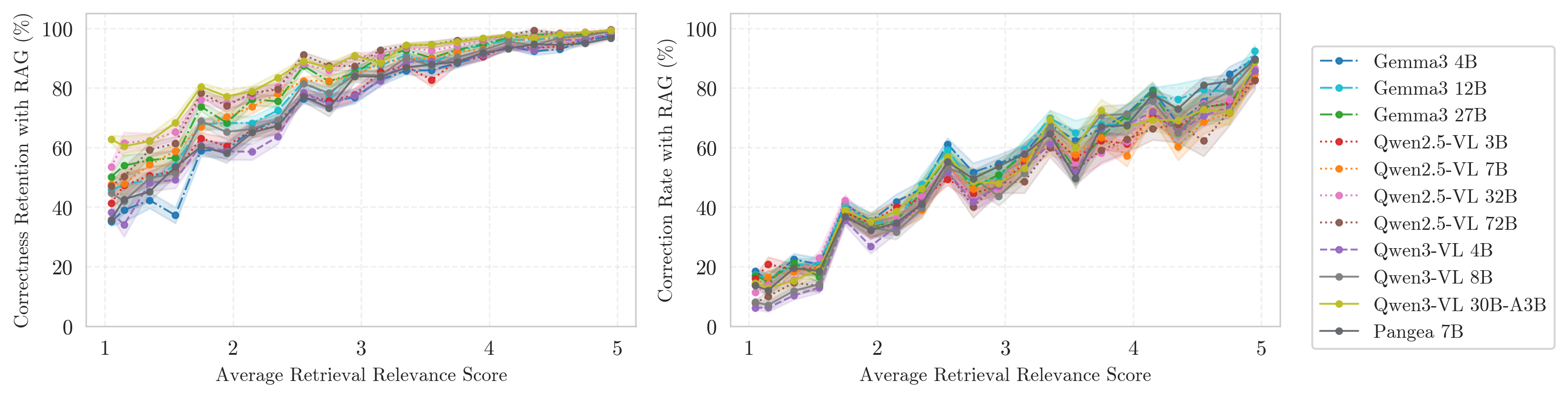}
  \caption{The effect of retrieval quality on RAG performance for various models on the \textsc{CVQA} dataset, using \texttt{mmE5} for multimodal retrieval. \textbf{Left:} The ``Correctness Retention'' rate measures the percentage of responses that were correct without RAG and remained correct with RAG. \textbf{Right:} The ``Correction Rate'' measures the percentage of responses that were incorrect without RAG but were successfully corrected by RAG.}
  \label{fig:ablation-cvqa-rag-compare-mme5}
\end{figure*}

\begin{figure*}[!th]
  \centering
  \includegraphics[width=\linewidth]{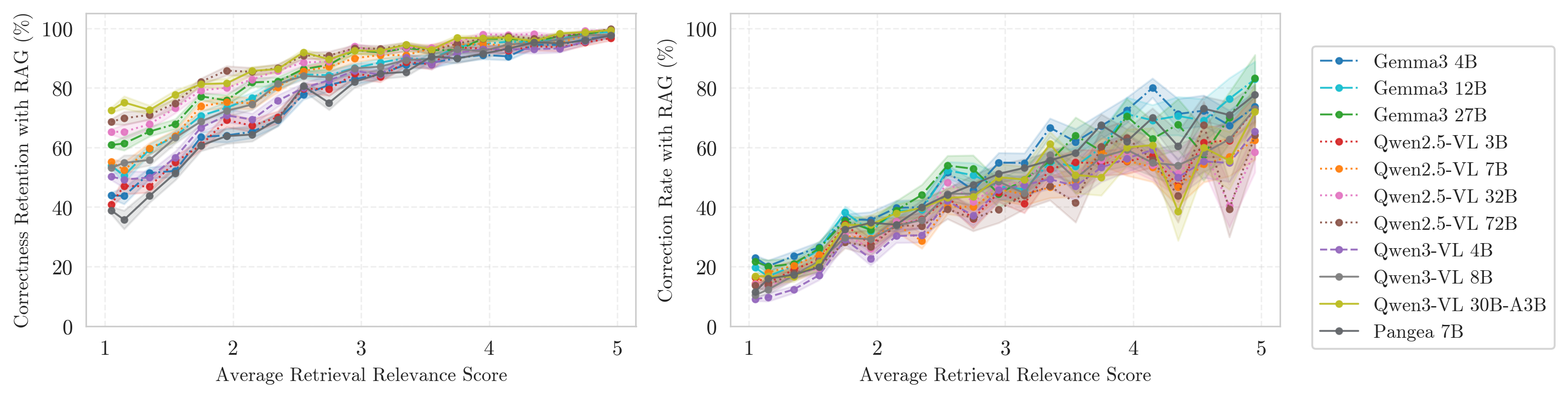}
  \caption{The effect of retrieval quality on RAG performance for various models on the \textsc{CVQA} dataset, using \texttt{B3} for multimodal retrieval. \textbf{Left:} The ``Correctness Retention'' rate measures the percentage of responses that were correct without RAG and remained correct with RAG. \textbf{Right:} The ``Correction Rate'' measures the percentage of responses that were incorrect without RAG but were successfully corrected by RAG.}
  \label{fig:ablation-cvqa-rag-compare-b3}
\end{figure*}

\subsection{Evaluation Metrics}
For VLM generations, we use macro-averaged accuracy for all datasets by comparing the multiple choice answer. For annotations we use VLM-as-a-judge using reasoning rubric based since it improves reasoning and more interpretable~\citep{lee2024prometheus, anugraha2025r3, anugraha2025mr3}. The rubrics and prompts for evaluation can be found in the Supplementary Materials.

\section{Results and Analysis}

\subsection{Overall Performance}

Figure~\ref{fig:overall-performance} presents the overall trends across all experiments. \texttt{Gemma3 27B} performs the best in both \textsc{CVQA} and \textsc{WorldCuisines} for baseline, with accuracy of 74.34\% and 66.20\%, respectively. As expected, providing the oracle context consistently yields the highest performance across all models and datasets, serving as an upper bound for the quality of contextual information. In this setting, \texttt{Gemma3 27B} performs the best in \textsc{CVQA}, while \texttt{Qwen-2.5-VL 72B} performs the best when given the oracle context in \textsc{WorldCuisines}.

Among retrieval strategies, text-based retrieval performs the worst, even worse than the baselines across model sizes and datasets, indicating that naively converting the image to text can introduce noise that harms VLM performance. In contrast, multimodal retrieval consistently outperforms text-based retrieval, although the \texttt{B3} embedding model shows comparatively lower gains. We also observe that reasoning VLMs consistently outperform non-reasoning models of comparable or larger size under RAG settings, suggesting that reasoning capability helps models better integrate retrieved context.

\begin{figure*}[!th]
  \centering
  \includegraphics[width=\linewidth]{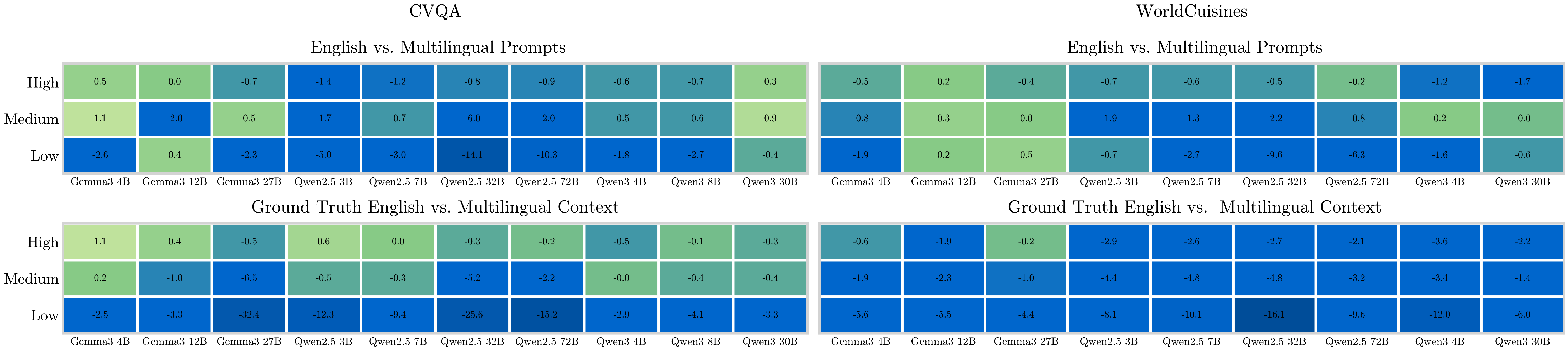}
  \caption{
Performance deltas (multilingual $-$ English) in $\methodname$ across languages grouped by vitality (high-, medium-, and low-resource). The top rows show the effect of switching from English to multilingual prompts, while the bottom rows show the effect of switching from English to multilingual oracle context. Negative values (darker green) indicate that the multilingual condition performs worse than English, while values near zero or positive (yellowish-green, hatched) indicate stability or gains.}
\label{fig:multi_perf}
\end{figure*}

\subsection{Model Scaling}

Figure~\ref{fig:overall-performance} illustrates how performance scales with model size across all families. While accuracy generally improves with scale, the benefit of retrieval does not follow the same trend. Text-based RAG performs the worst overall and scales poorly with model size, indicating that this retrieval approach is unlikely to provide meaningful benefits even for larger models.

For multimodal retrieval, \texttt{mmE5} and \texttt{B3} provide initial gains over the baseline, but these gains do not scale consistently. For larger models, the baseline eventually matches or surpasses multimodal RAG performance, suggesting that models struggle to effectively leverage retrieved context at scale. Reasoning models are more robust to this effect, maintaining retrieval gains longer as scale increases, but the overall trend holds where the slope of improvement diminishes with scale, and larger models show reduced reliance on external context regardless of retrieval strategy. This suggests a fundamental tension between model scale and retrieval utility, where stronger parametric knowledge increasingly competes with rather than complements externally retrieved evidence.

\subsection{When Does RAG Succeed and Fail?}

In order to understand the effect of retrieval quality on RAG performance, we analyze the quality of the retrieved context from both multimodal embeddings on both \textsc{CVQA} and \textsc{WorldCuisines} using VLM-as-a-judge, to see how relevant the retrieved context is with respect to the image, query, and the actual ground truth answer. We use two complementary metrics to support our analysis: \textit{correctness retention}, which is the percentage of initially correct answers that remain correct after RAG context is provided, and \textit{correction rate}, which is the percentage of initially incorrect answers successfully remediated by RAG context. Here, we will only show the plots for \textsc{CVQA} since \textsc{WorldCuisines} also provides the same linear trend, in which figures are shown in the supplementary materials instead. 

Figure~\ref{fig:ablation-cvqa-rag-compare-mme5} and Figure~\ref{fig:ablation-cvqa-rag-compare-b3} illustrate the impact of retrieval quality on the efficacy of RAG systems using \texttt{mmE5} and \texttt{B3}, respectively, applied to the \textsc{CVQA} dataset. Both metrics show a positive correlation between the average retrieval relevance score with the performance of all tested models. This demonstrates that the relevance of the retrieved context is aligned with the RAG system's overall success. However, their behavior differs in important ways. Correctness retention degrades sharply under poor retrieval, dropping to 40--60\% at scores below 2.0, confirming that irrelevant context actively misleads models into abandoning correct answers. As relevance improves, retention converges tightly across all models toward near-perfect rates (95--100\%), suggesting that high-quality retrieval reliably reinforces correct parametric knowledge regardless of model family or size.

The correction rate tells a different story. While high-relevance context enables models to fix 80--90\% of original errors, the correction rate never saturates to the same degree as retention, and model spread remains wide even at the highest relevance scores. This asymmetry indicates that leveraging retrieved evidence to overturn a wrong answer is fundamentally harder than preserving a correct one, and that current VLMs still struggle to reliably integrate externally retrieved evidence even when it is entirely correct. This gap is more pronounced with \texttt{B3} than \texttt{mmE5}, where wider inter-model variance suggests that weaker retrievers amplify differences in context integration ability across models.

Beyond the main trend, both plots reveal an important scaling effect. Larger models, exemplified by \texttt{Qwen2.5-VL 72B} and \texttt{Gemma3 27B}, exhibit greater reliance on their parametric knowledge. In the left plot, large models form the upper boundary of correctness retention (i.e., they more often preserve correct baseline answers), while in the right plot they frequently form the lower boundary of correction rate (i.e., they are less likely to change an incorrect baseline when given high-quality retrieved evidence). On the other hand, smaller model would take the counterparts, respectively on both multimodal embedding strategies. Taken together, these patterns indicate that model scale increases inertial priors, that is stronger internal beliefs that are less readily updated by external context. In other words, larger models show reduced context integration (or lower context susceptibility) as they are less prone to be misled by poor retrieval, but at the same time, also less likely to adopt corrective information supplied by good retrieval. This phenomenon suggests a potential point of diminishing returns for RAG, where beyond a certain scale, improvements in model capacity do not straightforwardly translate to better utilization of retrieved evidence.

\subsection{Multilingual Performance Gaps}

Our cross-lingual experiments reveal a strong English-centric bias in current VLMs. As shown in Figure~\ref{fig:multi_perf}, shifting from English to multilingual prompts consistently degrades performance across all resource levels, though the effect is relatively mild for high-resource languages (mostly within -1\% to -2\%) and becomes more severe for low-resource languages. This indicates that models best interpret task instructions in English regardless of the cultural context of the query.

The multilingual gap is far more pronounced even when the oracle context is provided in the target language rather than English. Contrary to the intuition that culturally aligned context should help, performance deteriorates dramatically, with drops as large as -32.4\% for \texttt{Qwen2.5-VL 32B} on \textsc{CVQA} and -28.8\% on \texttt{Pangea} on \textsc{CVQA} for low-resource languages. Note that \texttt{Pangea} is a model explicitly trained on multilingual and multicultural Wikipedia data, yet it is still among the most severely affected, suggesting that exposure to multilingual training data does not straightforwardly confer robustness to non-English retrieved context at inference time. This asymmetry between prompt switching and context switching indicates that models can tolerate non-English instructions to some extent, but fail much more severely when the retrieved evidence itself is in a non-English language, suggesting that cross-lingual evidence integration is a deeper bottleneck than instruction following.

This trend is not uniform across model families. The \texttt{Qwen} family exhibits a sharper collapse in low-resource settings compared to \texttt{Gemma}, showing that model scale alone does not resolve this bias. \texttt{Pangea} also shows some of the largest drops. An interesting observation is that smaller models show a lesser performance drop overall because they tend to code-switch to English even when prompted in a target language, whereas larger models attempt to respond fully in the target language and fail more dramatically as a result.

\section{Related Work}

Culture has been studied across multiple dimensions, including social norms~\cite{fung2023normsage}, country-specific variation~\cite{hu2024bridging,myung2024blend,romero2024cvqa}, general world knowledge~\cite{zhang2023miracl}, and food-related knowledge~\cite{ma2023food,agarwal2024indifoodvqa,hu2024bridging,li2024foodieqa,winata2025worldcuisines,yin2025foodlmm}. Understanding these facets is crucial for building AI systems that can reason appropriately across diverse cultural contexts. Early work on multicultural RAG has largely relied on machine-translated benchmarks~\cite{conia2024towards,lewis2020retrieval}. While this approach enables broader language coverage, it often fails to fully leverage high-quality human-curated resources and can introduce translation artifacts that obscure culturally specific nuances.

Several multilingual text-only retrieval datasets have been proposed, including MINERS~\cite{winata2024miners}, Mintaka~\cite{sen2022mintaka}, MIRACL~\cite{zhang2023miracl}, MKQA~\cite{longpre2021mkqa}, and MLQA~\cite{lewis2020mlqa}. These datasets span diverse domains such as books, geography, politics, and general knowledge, providing broad language coverage and enabling progress in multilingual natural language understanding. However, their focus on textual information limits their applicability to tasks that require multimodal reasoning~\cite{hudson2019gqa,tian2025core}. Tasks such as visual question answering~\cite{antol2015vqa,xia2025mmedrag,wu2025visual}, image-grounded reasoning~\cite{suhr2019nlvr,wang2025learningvisualgrounding,wu2025visual}, and culturally contextualized visual understanding remain largely unsupported. This gap highlights the need for benchmarks that jointly integrate multilingual and multimodal information~\cite{chen2023xvnli}. Recent efforts toward evaluating retrieval-augmented multimodal systems include MRAG-Bench~\cite{hu2025mragbench}, MRAMG-Bench~\cite{yu2025mramg}, MIRAGE-Bench~\cite{zhang2025mirage}, BordIRLines~\cite{li2024bordirlines}, and BERGEN~\cite{rau2024bergen}, but these benchmarks focus primarily on English settings.

At the same time, vision-language models (VLMs) have rapidly expanded their multilingual capabilities. Large proprietary or large-scale models such as Qwen2.5-VL~\cite{yang2024qwen2}, Qwen3-VL~\cite{bai2025qwen3}, Gemma3~\cite{team2025gemma}, and PaLI / PaLI-X~\cite{chen2023pali} demonstrate strong cross-modal reasoning across multiple languages. In parallel, a growing ecosystem of open multilingual VLMs has emerged, including InternVL~\cite{chen2024expanding}, mBLIP~\cite{geigle2024mblip}, PALO~\cite{maaz2024palo}, Maya~\cite{alam2024maya}, Aya Vision~\cite{dash2025aya}, and PaliGemma~\cite{beyer2024paligemma}. Despite these advances, systematic evaluation of multilingual VLMs on knowledge-intensive, culturally grounded multimodal tasks remains limited, highlighting the need for benchmarks that assess multilingual multimodal reasoning in realistic retrieval-augmented settings.

To address these gaps, we introduce $\methodname$ that combines multilingual and multimodal inputs, enabling end-to-end evaluation across languages with comprehensive experimental analysis.

\section{Conclusion}
In this work, we present $\methodname$, a massive-scale multilingual and multimodal benchmark spanning 42 languages, 56 regional dialects and registers, and over 80,000 culturally grounded image-question pairs for evaluating RAG in realistic settings. We conduct comprehensive experiments across 11 models and 3 retrieval configurations, including cross-lingual settings, to characterize the behavior of multilingual multimodal RAG systems. Our experiments reveal that while RAG reliably boosts smaller VLMs, larger models exhibit diminished correction rates and stronger reliance on parametric knowledge, and that multilingual context degrades performance even in models explicitly trained on multilingual data. These findings suggest that the fundamental challenge is not whether to retrieve, but how to enable effective integration of retrieved information, i.e. since larger models retain correct answers but fail to leverage retrieval for error correction, this may indicate misalignment between retrievers and foundation models. We therefore advocate for model-aware retrieval strategies that optimize for integration utility rather than query relevance alone, through directions such as joint retriever-VLM post-training or test-time adaptation. We hope $\methodname$ serves as a foundation for future work toward RAG systems that reason robustly across languages, modalities, and cultural contexts.

{
    \small
    \bibliographystyle{ieeenat_fullname}
    \bibliography{main}
}

\clearpage
\setcounter{page}{1}

\maketitlesupplementary

\section{Languages}
Table~\ref{tab:language_distribution} provides a comprehensive breakdown of the languages included in $\methodname$, which ensure rigorous evaluation of cross-lingual generalization. The languages include diverse language families (such as Indo-European, Sino-Tibetan, Afro-Asiatic, Austronesian, Japonic, Koreanic, Niger-Congo, Turkic, and Uralic) and varying resource levels. 

We categorize languages based on the taxonomy proposed by \citet{joshi2020state}, ranging from Class 0 to Class 5. This allows us to analyze how RAG performance correlates with the language vitality. Notably, our benchmark includes significant coverage of low-resource languages (Classes 0--2) such as Oromo, Tigrinya, Sundanese, and Sinhala, which are often underrepresented in standard VQA benchmarks.

 Unlike previous benchmarks that treat languages as monoliths, it $\methodname$ explicitly annotates regional dialects (e.g., Spanish across Spain, Argentina, Chile, Colombia, Ecuador, Mexico, and Uruguay) and social registers (e.g., formal vs. casual speech in Javanese, Korean, and Indonesian). This granularity is crucial for assessing cultural alignment, as the correct retrieval of cultural context often depends on recognizing dialect-specific nuances in the query.

\begin{figure*}[t!]
  \centering
  \includegraphics[width=0.95\linewidth]{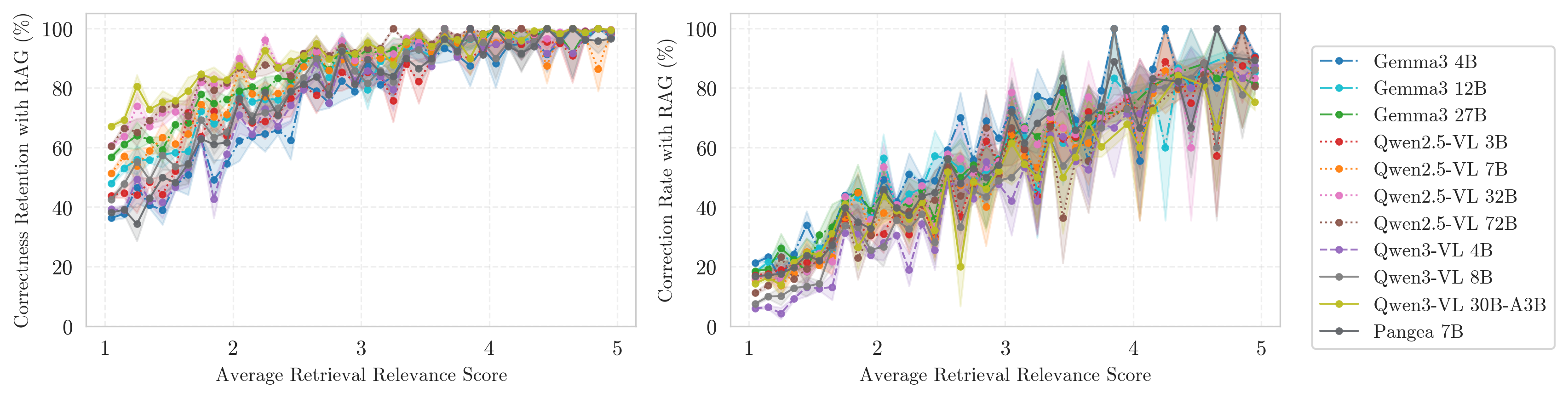}
  \caption{The effect of retrieval quality on RAG performance for various models on the \textsc{WorldCuisines} dataset, using mmE5 for multimodal retrieval.
\textbf{Left:} The ``Correctness Retention'' rate measures the percentage of responses that were correct without RAG and remained correct with
RAG. \textbf{Right:} The ``Correction Rate'' measures the percentage of responses that were incorrect without RAG but were successfully corrected
by RAG.}
  \label{fig:wc_lang_perf_effect}
\end{figure*}

\begin{figure*}[t!]
  \centering
  \includegraphics[width=0.95\linewidth]{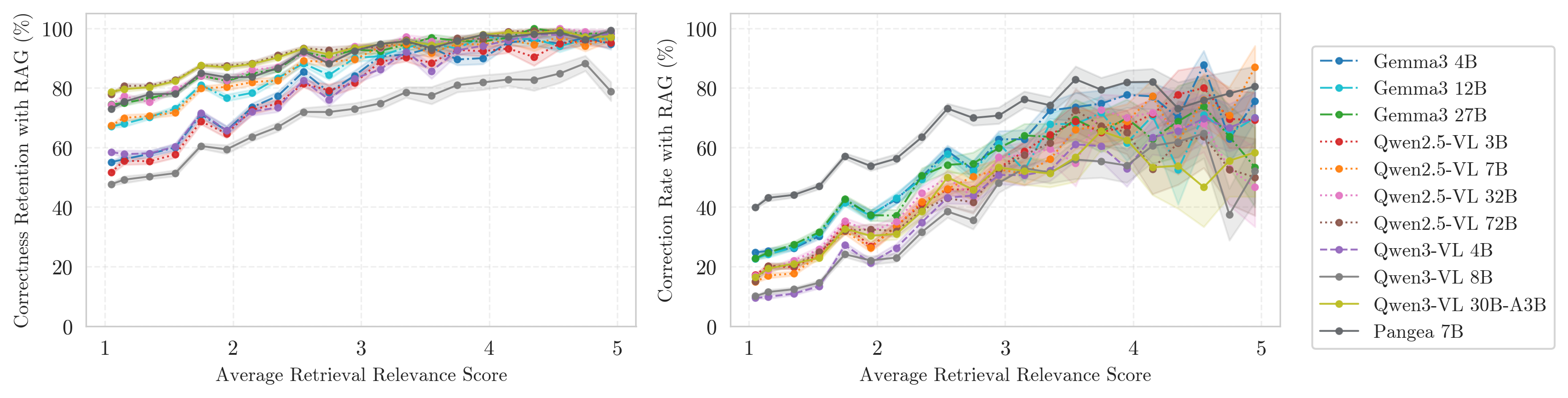}
  \caption{The effect of retrieval quality on RAG performance for various models on the \textsc{WorldCuisines} dataset, using B3 for multimodal retrieval.
\textbf{Left:} The ``Correctness Retention'' rate measures the percentage of responses that were correct without RAG and remained correct with
RAG. \textbf{Right:} The ``Correction Rate'' measures the percentage of responses that were incorrect without RAG but were successfully corrected
by RAG.}
  \label{fig:effect_wc_lang_perf}
\end{figure*}

\section{Human Evaluation}

\subsection{Human Verification on Generated Captions as Oracle Context for CVQA}

We use generated captions as oracle context because \textsc{CVQA} does not provide ground-truth evidence passages. The caption serves as a proxy for an upper bound on RAG performance. Note that the \textsc{CVQA} answers themselves are already human-annotated, and we simply generate the caption based on the images, questions, and the human-annotated answers themselves.

To further verify this, we conducted a human verification study by recruiting four annotators who evaluated 200 randomly sampled image-caption pairs on a 1--5 Likert scale for how well the caption supported answering the corresponding question. All samples received a score of 5 with full inter-annotator agreement, which is expected given the setup we have. For example, for a \textsc{CVQA} question asking "What part of the flag reflects the historical period?" given an image of the Romanian flag, the oracle context explicitly describes the central emblem as reflecting the historical period.

\subsection{Human Verification on VLM-as-a-judge}

We conduct a human validation study to examine whether our VLM-as-a-judge shows consistency with human scoring. Each of the \texttt{mmE5} and \texttt{B3} embedding models was evaluated on 100 samples, annotated by five human raters, and analyzed using five reliability metrics: Fleiss’ $\kappa$, Gwet’s AC2, Krippendorff’s $\alpha$, Conger’s $\kappa$, and Brennan-Prediger’s coefficient.

\begin{table}[!ht]
\centering
\begin{tabular}{lccc}
\toprule
\textbf{Metric} & \textbf{\texttt{mmE5}} & \textbf{\texttt{B3}} & \textbf{\texttt{All}} \\
\midrule
Fleiss’ $\kappa$ & 0.6573 & 0.4273 & 0.5488 \\
Gwet’s AC2 & 0.7225 & 0.5013 & 0.6179 \\
Krippendorff’s $\alpha$ & 0.6588 & 0.4300 & 0.5498 \\
Conger’s $\kappa$ & 0.6591 & 0.4432 & 0.5544 \\
Brennan-Prediger & 0.7115 & 0.4881 & 0.6059 \\
\bottomrule
\end{tabular}
\caption{Inter-rater agreement between human and model scores using different reliability metrics.}
\label{tab:human-eval}
\end{table}

Overall, the results indicate strong agreement for the \texttt{mmE5} model and moderate agreement for the \texttt{B3} model. This suggests that the lower retrieval performance observed for \texttt{B3} may stem from an understanding mismatch, where specific chunks receive higher localized scores despite inconsistent overall perception.

\section{Detailed Results}

In this section, we provide a granular analysis of the performance metrics reported in Tables~\ref{tab:wc-results} and~\ref{tab:cvqa-results}. We focus on the interaction between model scale, retrieval modality, and the upper bounds established by oracle contexts.

\subsection{Inverse Scaling of Retrieval Benefits}

A central finding in our experiments is the inverse correlation between model parameter count and the relative performance gain provided by RAG. 

\paragraph{Small Models ($<$14B).} As shown in Table~\ref{tab:cvqa-results}, smaller models exhibit substantial gains from multimodal retrieval. For instance, on the \textsc{CVQA} benchmark, \texttt{Gemma3 4B} improves from a baseline of 59.22\% to 64.96\% (+5.74\%) when using \texttt{mmE5} retrieval. Similarly, \texttt{Qwen2.5-VL 3B} sees an improvement of +7.34\%. This suggests that smaller models, which lack extensive parametric knowledge, rely heavily on retrieved context to ground their answers.

\paragraph{Large Models ($>$14B).} Conversely, larger models show diminishing returns or performance degradation. \texttt{Gemma3 27B} on \textsc{CVQA} regresses from 74.34\% (Baseline) to 72.59\% with \texttt{mmE5} RAG. \texttt{Qwen2.5-VL 72B} exhibits a similar pattern. This implies that for large models, imperfect retrieval acts as a distractor rather than an aid; the model's internal parametric knowledge is often more accurate than the noisy context retrieved.

\label{sec:supp_languages}

\begin{table*}[!ht]
\centering
\resizebox{.9\textwidth}{!}{
    \begin{tabular}{l|cc|cc|cccc}
    \toprule
    \multirow{2}{*}{\textbf{Model}} & \multicolumn{2}{c|}{\textbf{No RAG}} & \multicolumn{2}{c|}{\textbf{Oracle Context}} & \multicolumn{4}{c}{\textbf{RAG}} \\
    & Baseline & + Multilingual Prompt & Eng. & Multilingual & Eng. Cap. & Oracle Eng. & mmE5 & B3 \\
    \midrule
    Gemma3 4B & 59.22 & 59.32 & 95.01 & 94.50 & 53.16 & 82.02 & 64.96 & 56.71 \\
    Gemma3 12B & 69.43 & 69.43 & 98.09 & 97.31 & 61.50 & 85.33 & 69.99 & 63.05 \\
    Gemma3 27B & 74.34 & 73.89 & 98.61 & 92.13 & 66.04 & 86.86 & 72.59 & 68.03 \\
    \midrule
    Qwen2.5-VL 3B & 56.29 & 55.09 & 93.97 & 91.59 & 52.63 & 79.68 & 63.63 & 52.85 \\
    Qwen2.5-VL 7B & 62.26 & 61.47 & 95.32 & 93.46 & 59.26 & 82.17 & 67.05 & 59.04 \\
    Qwen2.5-VL 32B & 68.75 & 65.37 & 97.14 & 92.12 & 65.44 & 85.88 & 71.72 & 65.49 \\
    Qwen2.5-VL 72B & 73.51 & 71.19 & 97.48 & 94.52 & 68.38 & 86.23 & 72.03 & 68.73 \\
    \midrule
    Qwen3-VL 4B Think & 58.48 & 57.88 & 94.65 & 93.94 & 50.95 & 78.97 & 62.00 & 53.28 \\
    Qwen3-VL 8B Think & 64.10 & 63.54 & 96.25 & 95.36 & 55.95 & 82.10 & 66.21 & 58.33 \\
    Qwen3-VL 30B A3B Think & 72.34 & 72.35 & 97.51 & 96.72 & 68.82 & 87.14 & 74.38 & 69.80 \\
    \midrule
    Pangea 7B & 48.99 & 45.45 & 94.33 & 87.94 & 46.86 & 78.63 & 61.93 & 50.11 \\
    \bottomrule
    \end{tabular}
}
\caption{Detailed results for \textsc{CVQA} across different multilingual settings and RAG settings.}
\label{tab:cvqa-results}
\end{table*}

\begin{table*}[t!]
\centering
\resizebox{.9\textwidth}{!}{
    \begin{tabular}{l|cc|cc|cccc}
    \toprule
    \multirow{2}{*}{\textbf{Model}} & \multicolumn{2}{c|}{\textbf{No RAG}} & \multicolumn{2}{c|}{\textbf{Oracle Context}} & \multicolumn{4}{c}{\textbf{RAG}} \\
    & Baseline & + Multilingual Prompt & Eng. & Multilingual & Eng. Cap. & Oracle Eng. & mmE5 & B3 \\
    \midrule
    Gemma3 4B & 48.26 & 47.22 & 57.19 & 54.39 & 39.60 & 47.91 & 52.73 & 47.20 \\
    Gemma3 12B & 62.46 & 62.71 & 74.24 & 70.97 & 49.08 & 56.76 & 59.45 & 57.25 \\
    Gemma3 27B & 66.20 & 66.24 & 78.43 & 76.50 & 55.56 & 62.70 & 63.83 & 62.66 \\
    \midrule
    Qwen2.5-VL 3B & 46.22 & 44.95 & 57.27 & 52.07 & 39.43 & 46.38 & 51.08 & 41.49 \\
    Qwen2.5-VL 7B & 53.87 & 52.32 & 64.22 & 58.28 & 47.96 & 55.08 & 56.02 & 50.08 \\
    Qwen2.5-VL 32B & 60.00 & 55.75 & 74.31 & 66.35 & 53.39 & 61.94 & 62.89 & 57.48 \\
    Qwen2.5-VL 72B & 65.14 & 62.67 & 79.68 & 74.64 & 58.03 & 65.95 & 63.68 & 61.76 \\
    \midrule
    Qwen3-VL 4B Think & 47.22 & 46.34 & 59.39 & 52.93 & 34.86 & 44.37 & 45.93 & 39.29 \\
    Qwen3-VL 8B Think & 53.79 & 52.70 & 68.05 & 61.93 & 40.84 & 49.69 & 51.09 & 42.42 \\
    Qwen3-VL 30B A3B Think & 65.54 & 64.77 & 77.61 & 74.35 & 59.69 & 66.00 & 65.68 & 62.26 \\
    \midrule
    Pangea 7B & 47.05 & 36.53 & 61.80 & 47.32 & 35.88 & 44.68 & 50.99 & 40.54 \\
    \bottomrule
    \end{tabular}
}
\caption{Detailed results for \textsc{WorldCuisines} across different multilingual settings and RAG settings.}
\label{tab:wc-results}
\end{table*}

\subsection{Oracle-RAG Performance Gap}

\begin{figure*}[!ht]
  \centering
  \begin{subfigure}[t]{.85\linewidth}
    \centering
    \includegraphics[width=\linewidth]{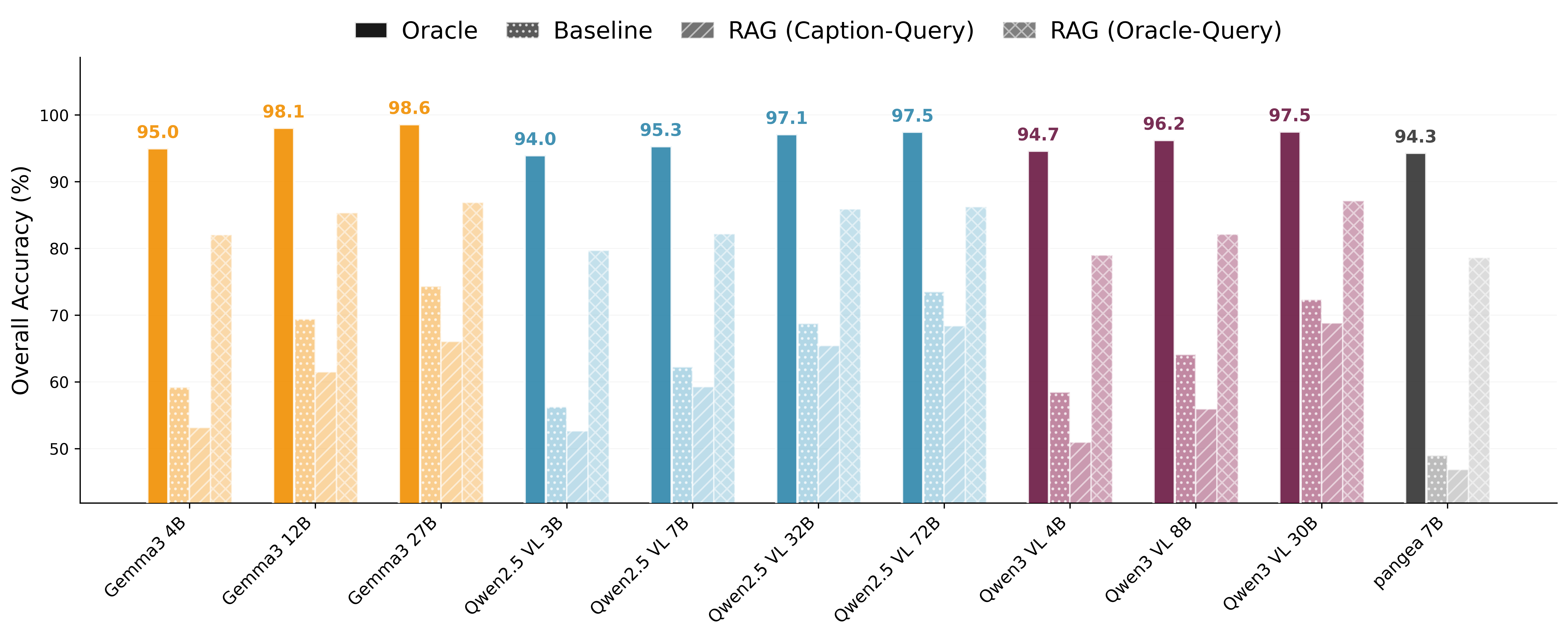}
    \caption{\textsc{CVQA}.}
    \label{fig:oracle_cvqa}
  \end{subfigure}
  
  \vspace{0.3cm}
  
  \begin{subfigure}[t]{.85\linewidth}
    \centering
    \includegraphics[width=\linewidth]{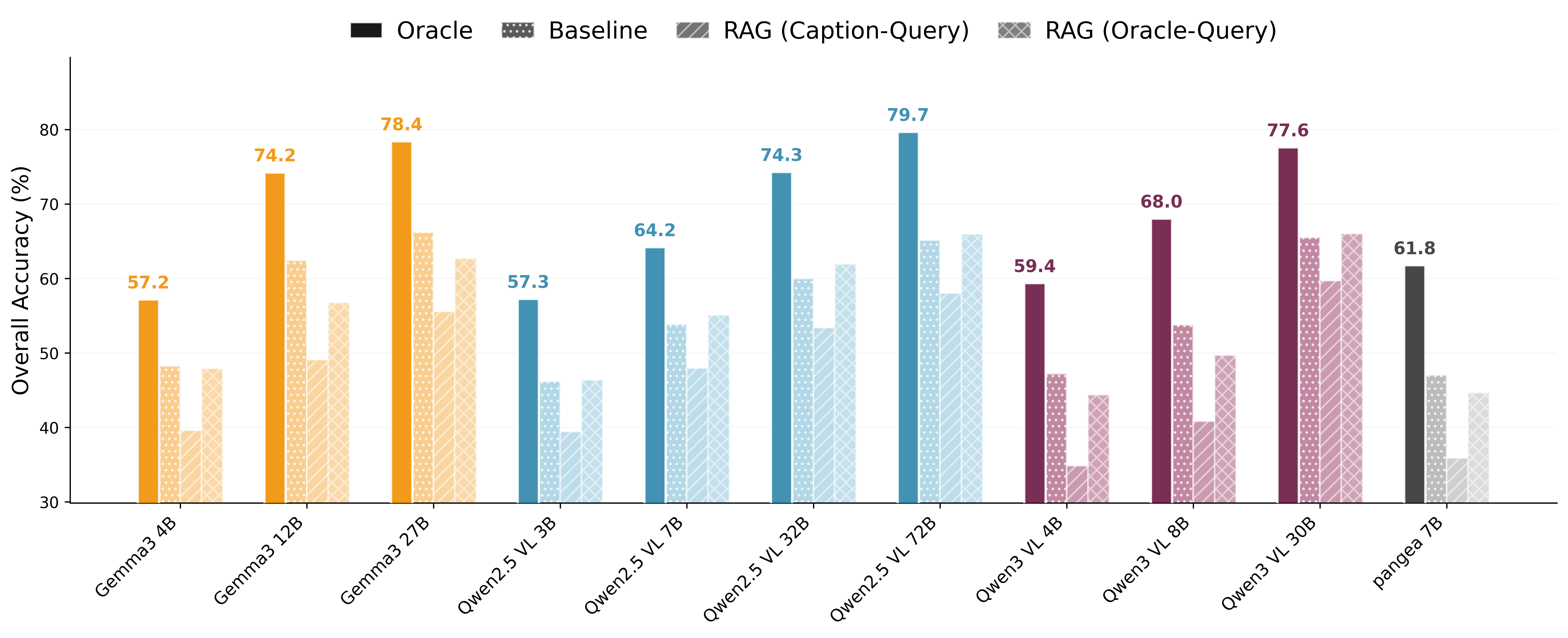}
    \caption{\textsc{WorldCuisines}.}
    \label{fig:oracle_wc}
  \end{subfigure}
  \caption{Comparison of oracle context versus RAG performance across model families on \textbf{(a)} \textsc{CVQA} and \textbf{(b)} \textsc{WorldCuisines}. The performance gap widens with model scale, indicating that while larger VLMs can effectively leverage perfect context, current retrieval systems fail to provide evidence of sufficient quality to match oracle performance.}
  \label{fig:oracle_vs_rag}
\end{figure*}

Figures~\ref{fig:oracle_cvqa} and ~\ref{fig:oracle_wc} illustrate the substantial performance gap between providing oracle context and retrieval-augmented generation across both benchmarks. On \textsc{CVQA} (Figure~\ref{fig:oracle_cvqa}), oracle context consistently achieves 94--99\% accuracy across all models, establishing a clear upper bound. In contrast, even the best RAG configurations using multimodal retrieval (\texttt{mmE5} or Oracle-Query RAG) achieve only 64--74\% accuracy for the largest models, revealing a gap of 20--30\%. This disparity is even more pronounced on \textsc{WorldCuisines} (Figure~\ref{fig:oracle_wc}), where oracle performance reaches 74--80\%, while RAG variants plateau at 62--68\%. The caption-based RAG approach consistently underperforms, often falling below the baseline. Notably, the gap between oracle and RAG widens as model size increases, indicating that while larger models can effectively leverage perfect context, they struggle to extract useful information from imperfect retrieval. This underscores that current retrieval systems are far from providing the quality of evidence that VLMs can utilize, hence pointing to retrieval quality as the primary bottleneck in multilingual multimodal RAG pipelines.

\begin{figure*}[t!]
  \centering
  \includegraphics[width=\linewidth]{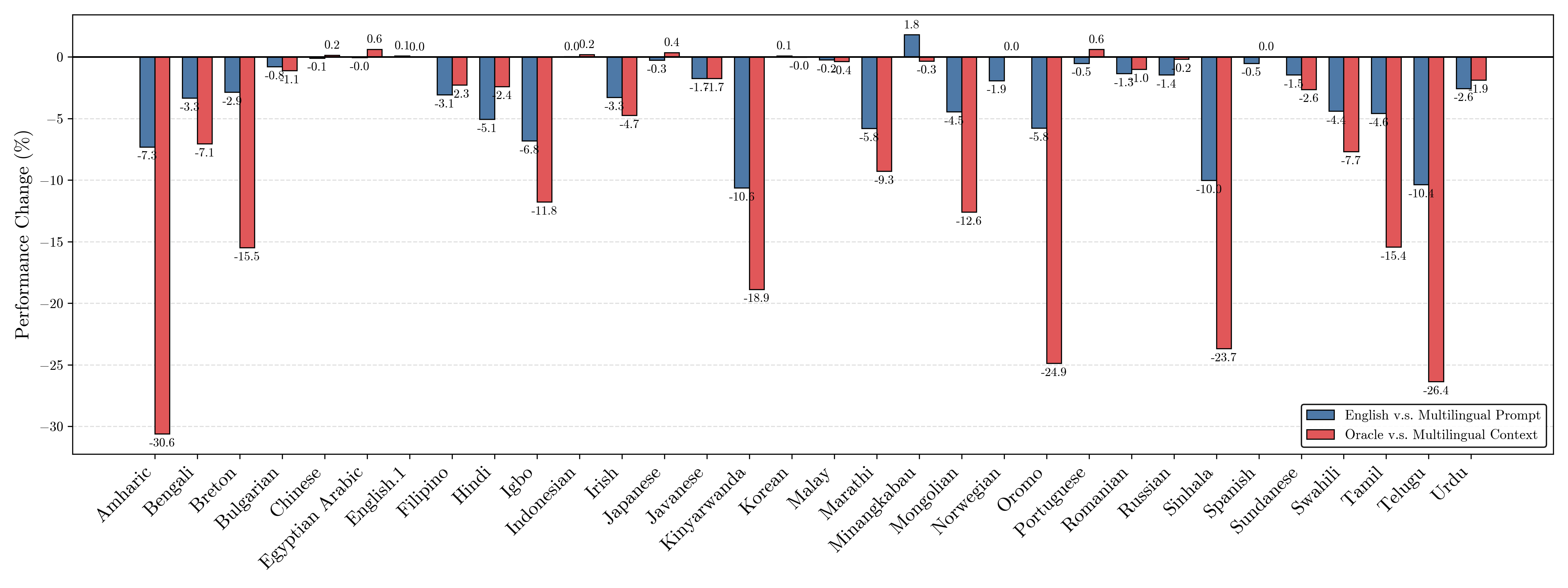}
  \caption{Language-wise performance change on \textsc{CVQA} when switching from English to multilingual prompts. Similar to \textsc{WorldCuisines}, low-resource languages exhibit substantial performance degradation.}
  \label{fig:cvqa_lang_perf}
\end{figure*}

\begin{figure*}[!t]
  \centering
  \includegraphics[width=\linewidth]{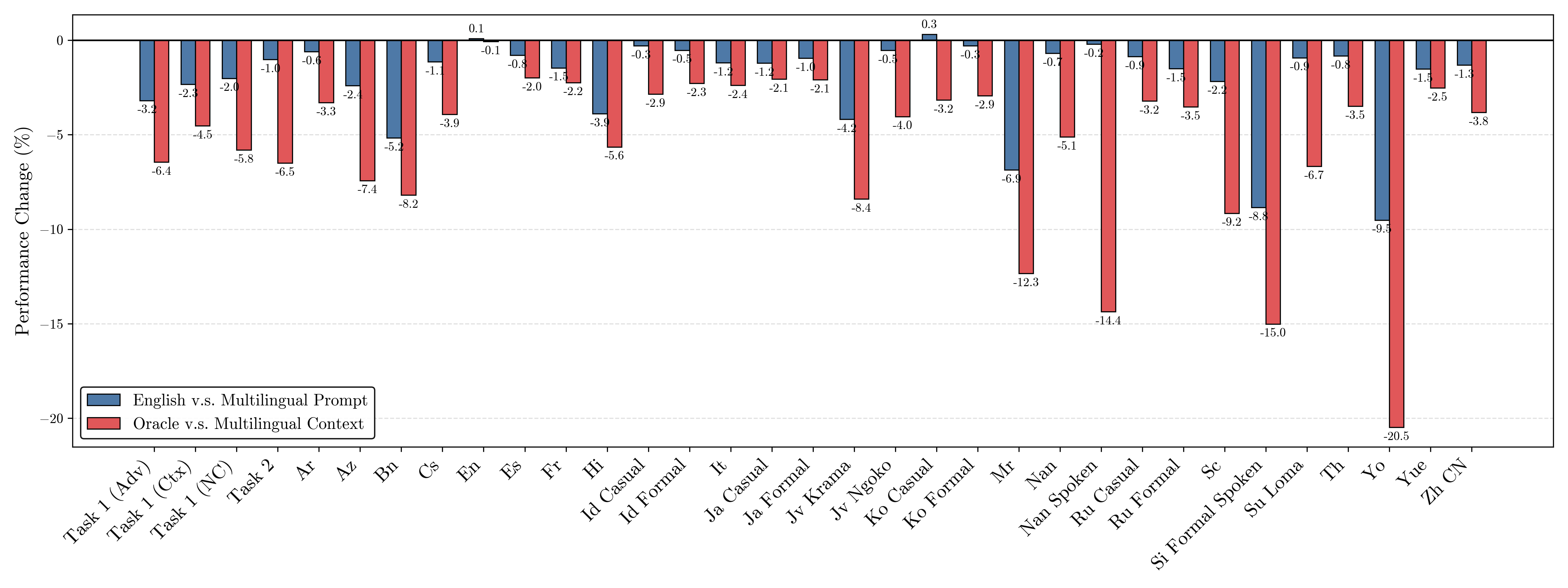}
  \caption{Language-wise performance change on \textsc{WorldCuisines} when switching from English to multilingual prompts. Negative values indicate performance degradation, with low-resource languages showing the most significant drops.}
  \label{fig:wc_lang_perf}
\end{figure*}

\subsection{Language-Wise Performance Analysis}

To further investigate the multilingual capabilities of current VLMs, Figures~\ref{fig:wc_lang_perf} (\textsc{WorldCuisines}) and \ref{fig:cvqa_lang_perf} (\textsc{CVQA}) break down the performance impact of language choice on instructions. These figures represent the performance change when switching from English instructions to the target language. We observe similar patterns across both benchmarks. While high-resource languages like Chinese, Spanish, and French maintain relatively same performance, low-resource languages such as Amharic, Telugu, and Oromo suffer significant degradation, often dropping by over 5--10\%. This confirms an inherent bias in current instruction-tuning approaches: despite being capable of generating multilingual text, these models follow reasoning instructions significantly better when presented in English.

Furthermore, figures reveal a critical limitation regarding contextual grounding. Intuitively, one might expect that answering a culture-specific question would be easier when the supporting evidence (oracle context) is provided in that culture's native language. However, our results indicate the opposite. Across the majority of languages, particularly in the \textsc{WorldCuisines} for languages like Yoruba and Marathi, providing oracle context in the target language causes a sharper performance decline than simply changing the prompt language. This inverse effect indicates that VLMs treat English as a reasoning pivot: they struggle to integrate non-English evidence, preferring English context even for culture-specific queries where native-language grounding should be advantageous.

\section{Prompts}
\label{sec:prompts}

\subsection{Translation Prompts}
\label{sec:translation_prompts}

To generate the multilingual instructional prompt, we utilized an LLM with the prompt structure shown in Figure~\ref{fig:translation_prompt}. The prompt is designed to ensure the translation maintains the specific formatting required for template substitution (e.g., preserving double curly braces).

\begin{figure*}[!ht] 
    \begin{lstlisting}[
        basicstyle=\footnotesize\ttfamily, 
        breaklines=true,           
        frame=single,             
        showstringspaces=false,    
        columns=fullflexible        
    ]
You are an expert translator with specialization in prompt engineering. Your task is to translate the string values of the following JSON object into {target_language}. 

### Guidelines:
1. Tone & Style: The text is used to prompt an AI model. Ensure the translation is clear, concise, and instructional. It should sound natural and culturally appropriate for a native speaker of {target_language}, but maintain the directive nature of the original text.
2. Placeholders: Do NOT translate or alter any text inside curly braces (e.g., keep '{{input}}' or '{{name}}' exactly as they are).
3. Structure: Keep the JSON keys exactly the same. Only translate the values.

### Output Format:
Return ONLY the raw JSON string. 
    - Do NOT use Markdown code blocks (no ```json).
    - Do NOT add explanations or conversational text.
    - Ensure the output is valid, parseable JSON.

### Input JSON:
{input_json_string}
    \end{lstlisting}
    \caption{Prompt for translating system instructions to target language.}
    \label{fig:translation_prompt}
\end{figure*}

\subsection{Evaluation Prompts}
\label{sec:eval_prompts}

To assess performance and quality, we utilized two distinct prompts. The first is a ``VLM-as-a-judge'' prompt used to evaluate the relevance of retrieved context (Figure~\ref{fig:vlm_judge_prompt}). The second is the inference prompt used to generate the final multiple-choice answer given the context (Figure~\ref{fig:inference_prompt}).

\begin{figure*}[!ht] 
    \begin{lstlisting}[basicstyle=\footnotesize\ttfamily, breaklines=true, frame=single, showstringspaces=false, columns=fullflexible]
You are an expert evaluator for a Vision-Language RAG system. Given an image and a question, assess how well the provided textual context supports answering the image-based question, considering both its relevance to the question and its helpfulness in reaching or verifying the ground truth answer. You must evaluate the context according to the given rubric by providing a short explanation for your reasoning and then assign a single holistic score (1-5).

### Question
{{ question }}

### Ground Truth Answer
{{ ground_truth_answer }}

### Context
{{ context }}

### Evaluation Rubric
1: The context is completely irrelevant or misleading as the context provides no useful information for answering the question.
2: The context is slightly related but mostly unhelpful as the context contains minimal connection or value toward the answer.
3: The context is somewhat relevant and partially useful as the context offers limited insight or indirect clues toward the answer.
4: The context is mostly relevant and helpful as the context supports reasoning toward the correct answer though not perfectly comprehensive.
5: The context is highly relevant and directly helpful as the context clearly supports or confirms the correct ground truth answer.

### Response Format
Provide your response in the following JSON format:

{{ format | schema }}

### Response
    \end{lstlisting}
    \caption{Prompt for evaluating the relevance of retrieved context (VLM-as-a-judge).}
    \label{fig:vlm_judge_prompt}
\end{figure*}

\subsection{Inference Prompts}
\label{sec:inf_prompts}

To generate the final answer for the visual question answering task, we employ the structured inference prompt displayed in Figure~\ref{fig:inference_prompt}. This prompt aggregates the input question, the retrieved context passages (if available), and the multiple-choice options. The model is instructed to reason based on the provided context and output the answer in a strict JSON format to facilitate automated parsing.

\begin{figure*}[!ht] 
    \begin{lstlisting}[basicstyle=\footnotesize\ttfamily, breaklines=true, frame=single, showstringspaces=false, columns=fullflexible]
Given the multiple-choice question below, choose the single best answer based on the question and any relevant context provided. Respond only with the number of the correct option (i.e., 1, 2, 3, or 4). Use the context if helpful, but ignore unrelated information.

### Question
{{ question }}
{% if context_list %}

### Context
{% for context in context_list %}
- {{ context }}
{% endfor %}

{% endif %}

### Options
{% for option in options %}
{{ loop.index }}. {{ option }}
{% endfor %}

### Answer Format
Provide your response in the following JSON format:

{{ format | schema }}

### Response
    \end{lstlisting}
    \caption{Prompt template for the multiple-choice VQA task with retrieval augmentation.}
    \label{fig:inference_prompt}
\end{figure*}

\begin{table*}[t!]
    \centering
    \resizebox{.9\textwidth}{!}{
    \begin{tabular}{lcccccc}
        \toprule
        \textbf{Language} & \textbf{Family} & \textbf{Resource Class}$^\dagger$ & \textbf{Register} & \textbf{Regional Dialects} & \textbf{In \textsc{CVQA}} & \textbf{In \textsc{WorldCuisines}} \\
        \midrule
        Amharic & Afro-Asiatic & 2 & & Ethiopia & \checkmark & \\
        Arabic & Afro-Asiatic & 5 & & Arab & & \checkmark \\
        Azerbaijani & Turkic & 1 & & & & \checkmark \\
        Bengali & Indo-European & 3 & & India & \checkmark & \checkmark \\
        Breton & Indo-European & 1 & & France & \checkmark & \\
        Bulgarian & Indo-European & 3 & & Bulgaria & \checkmark & \\
        Cantonese & Sino-Tibetan & 1 & & & & \checkmark \\
        Chinese & Sino-Tibetan & 5 & & China & \checkmark & \checkmark \\
        Chinese & Sino-Tibetan & 5 & & Singapore & \checkmark & \checkmark \\
        Czech & Indo-European & 4 & & & & \checkmark \\
        Egyptian Arabic & Afro-Asiatic & 3 & & Egypt & \checkmark & \\
        English & Indo-European & 5 & & United States & \checkmark & \checkmark \\
        French & Indo-European & 5 & & France & & \checkmark \\
        Hokkien & Sino-Tibetan & 0 & Written & Medan & & \checkmark \\
        Hokkien & Sino-Tibetan & 0 & Spoken & Medan & & \checkmark \\
        Hindi & Indo-European & 4 & & India & \checkmark & \checkmark \\
        Igbo & Niger-Congo & 1 & & Nigeria & \checkmark & \\
        Indonesian & Austronesian & 3 & Formal & Indonesia & \checkmark & \checkmark \\
        Indonesian & Austronesian & 3 & Casual & Indonesia & & \checkmark \\
        Irish & Indo-European & 2 & & Ireland & \checkmark & \\
        Italian & Indo-European & 4 & & & & \checkmark \\
        Japanese & Japonic & 5 & Formal & Japan & \checkmark & \checkmark \\
        Japanese & Japonic & 5 & Casual & Japan & & \checkmark \\
        Javanese & Austronesian & 1 & Krama & Java & \checkmark & \checkmark \\
        Javanese & Austronesian & 1 & Ngoko & Java & & \checkmark \\
        Kinyarwanda & Niger-Congo & 1 & & Rwanda & \checkmark & \\
        Korean & Koreanic & 4 & Formal & South Korea & \checkmark & \checkmark \\
        Korean & Koreanic & 4 & Casual & South Korea & & \checkmark \\
        Marathi & Indo-European & 2 & & India & \checkmark & \\
        Malay & Austronesian & 3 &  & Malaysia & \checkmark & \\
        Minangkabau & Austronesian & 1 & & Indonesia & \checkmark & \\
        Mongolian & Mongolic & 1 &  & Mongolia & \checkmark & \\
        Norwegian & Indo-European & 1 & & Norway & \checkmark & \\
        Oromo & Afro-Asiatic & 1 & & Ethopia & \checkmark & \\
        Portuguese & Indo-European & 4 & & Brazil & \checkmark & \\
        Romanian & Indo-European & 3 & & Romania & \checkmark & \\
        Russian & Indo-European & 5 & Formal & Russia & \checkmark & \checkmark \\
        Russian & Indo-European & 5 & Casual & Russia & & \checkmark \\
        Sardinian & Indo-European & 1 & & Italy & & \checkmark \\
        Sinhala & Indo-European & 0 & Formal & Sri-Lanka & \checkmark & \checkmark \\
        Spanish  & Indo-European & 5 &  & Spain & \checkmark & \checkmark \\
        Spanish  & Indo-European & 5 &  & Argentina & \checkmark & \\
        Spanish  & Indo-European & 5 &  & Chile & \checkmark & \\
        Spanish  & Indo-European & 5 &  & Colombia & \checkmark & \\
        Spanish  & Indo-European & 5 &  & Ecuador & \checkmark & \\
        Spanish  & Indo-European & 5 &  & Mexico & \checkmark & \\
        Spanish  & Indo-European & 5 &  & Uruguay & \checkmark & \\
        Sundanese & Austronesian & 1 & Loma & Indonesia & \checkmark & \checkmark \\
        Swahili & Niger-Congo & 2 & & Kenya & \checkmark & \\
        Tagalog & Austronesian & 3 & & Phillipines & \checkmark & \checkmark \\
        Tamil & Indo-European & 3 & & India & \checkmark & \\
        Telugu & Indo-European & 1 & & India & \checkmark & \\
        Thai & Kra-Dai & 3 & & & & \checkmark \\
        Urdu & Indo-European & 3 & & India & \checkmark & \\
        Urdu & Indo-European & 3 & & Pakistan & \checkmark & \\
        Yoruba & Niger-Congo & 2 & & & & \checkmark \\
        \bottomrule
    \end{tabular}
    }
    \caption{Languages used in $\methodname$. Resource classes follow the 0--5 scale of \citet{joshi2020state}.}
    \label{tab:language_distribution}
\end{table*}

\section{Hyper-parameters}
For all inference runs, we use 4 NVIDIA H100 80GB GPUs with vLLM and set the maximum output length to 16,384 tokens. For \texttt{Qwen3-VL}, we use the recommended generation settings: temperature = 1.0, presence penalty = 0.0, repetition penalty = 1.0, top-k = 20, and top-p = 0.95. For \texttt{Gemma3}, we use top-k = 64 and top-p = 0.95. For \texttt{Qwen2.5-VL} and \texttt{Pangea}, we follow the recommended settings of repetition penalty = 1.05 and temperature = 0.

\end{document}